%% file: main.tex
\documentclass[10pt,twocolumn,letterpaper]{article}

\usepackage[]{cvpr}         

\input{preamble}

\definecolor{cvprblue}{rgb}{0.21,0.49,0.74}
\definecolor{darkred}{RGB}{200, 0, 0}
\definecolor{darkyellow}{RGB}{250, 153, 0}
\definecolor{darkgreen}{RGB}{0, 180, 0}
\definecolor{lise}{RGB}{128,0,32}
\definecolor{deepblue}{RGB}{68,85,102}
\definecolor{graygreen}{RGB}{85,107,47}
\definecolor{tomato}{RGB}{255,99,71}
\definecolor{india}{RGB}{205,92,92}

\definecolor{saddlebrown}{RGB}{139, 69, 19}

\definecolor{forestgreen}{RGB}{34, 139, 34}
\definecolor{maroon}{RGB}{217,38,38}
\definecolor{navy}{RGB}{164, 101, 138}
\definecolor{moccasin}{RGB}{230, 159, 0}

\definecolor{darkred}{RGB}{200, 0, 0}
\definecolor{darkyellow}{RGB}{250, 153, 0}
\definecolor{darkgreen}{RGB}{0, 180, 0}
\usepackage{adjustbox}
\usepackage{wrapfig}
\usepackage[pagebackref,breaklinks,colorlinks,allcolors=cvprblue]{hyperref}

\def\confName{CVPR}
\def\confYear{2027}

\title{EG-Stereo: Geometric Edge Guidance for Stereo Matching at Depth Boundaries}

\author{
Qi Xiong$^{1}$  \quad Jiyuan Ye$^{1}$ \quad Qinghua Zhao$^{2, *}$ \\[2pt]
$^{1}$Xidian University \quad $^{2}$Hefei University\\
{\tt\small \{xiongqi, zzyeji\}@stu.xidian.edu.cn, \ zhaoqh@hfuu.edu.cn}
}

\begin{document}
\maketitle

\input{sec/0_abstract}

\input{sec/1_intro}

\input{sec/2_related}

\input{sec/3_method}

\input{sec/4_experiments}

\input{sec/5_conclusion}

{
    \small
    \bibliographystyle{ieeenat_fullname}
    \bibliography{main}
}


\end{document}

%% file: preamble.tex
\usepackage{dsfont}     
\usepackage{array}      
\usepackage{multirow}   
\usepackage{colortbl}   
\definecolor{firstbg}{HTML}{FFBFBF}   
\definecolor{secondbg}{HTML}{FFFFBF}  



%% file: sec/0_abstract.tex
\begin{abstract}
Depth boundaries remain a major source of error in stereo matching. For IGEV, a widely used stereo matching baseline, on Scene Flow, geometric edge pixels account for only $6\%$ of evaluated pixels but about half of the total absolute disparity error. Photometric edges can respond to texture changes without corresponding depth discontinuities, making them unreliable cues for boundary refinement.  We propose EG-Stereo, which learns a geometric edge map from disparity supervision and uses it as a shared spatial cue for disparity refinement and training. The predicted map conditions full-resolution residual correction alongside the backbone disparity and image details, and increases disparity error penalties at predicted boundaries. Joint optimization retains geometric edge supervision while allowing the edge representation to adapt to disparity correction. On the Scene Flow dataset, EG-Stereo reduces end-point error (EPE) at depth boundaries by $9.5\% \sim14.3\%$ across RAFT-Stereo, IGEV, and MonSter, while also reducing overall EPE. After fine-tuning, our IGEV-based model improves upon IGEV on all fourteen reported metrics across ETH3D, Middlebury 2014, KITTI 2012, and KITTI 2015. The source code will be made publicly available upon acceptance.
\end{abstract}

%% file: sec/1_intro.tex
\section{Introduction}
\label{sec:intro}

Stereo matching estimates disparity by finding pixel correspondences between rectified image pairs of the same scene, providing the geometric basis for recovering scene depth. It supports applications such as autonomous driving and 3D reconstruction~\cite{Geiger2012CVPR,Menze2015ISA}.

\begin{figure}[t]
  \centering
  \begin{subfigure}{\linewidth}
    \centering
    \includegraphics[width=\linewidth]{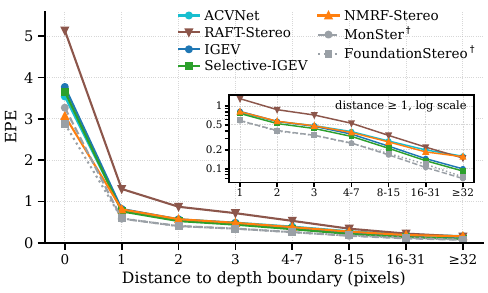}
    \caption{EPE versus distance to depth boundaries.}
    \label{fig:teaser_distance}
  \end{subfigure}


  \begin{subfigure}{\linewidth}
    \centering
    \includegraphics[width=\linewidth]{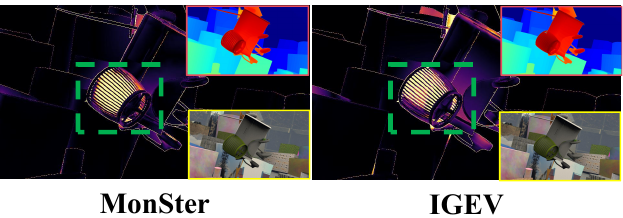}
    \caption{Disparity error maps of MonSter and IGEV.}
    \label{fig:teaser_demo}
  \end{subfigure}
  \caption{Disparity errors at depth boundaries.}
  \label{fig:teaser}
\end{figure}

Recent stereo methods have improved disparity estimation~\cite{RAFT-Stereo,CREStereo,IGEV,MonSter,DEFOM-Stereo,BridgeDepth}. Previous work has addressed boundary accuracy through image-guided disparity refinement~\cite{StereoNet}, joint stereo matching and edge detection~\cite{EdgeStereo}, and supervision that accounts for multiple disparity modes at edge pixels~\cite{ADL}. Nevertheless, our evaluation of seven existing stereo methods on Scene Flow~\cite{MIFDB16} shows that disparity errors remain higher at depth boundaries than in non-boundary regions. \Cref{fig:teaser}(a) shows mean end-point error (EPE) as a function of pixel distance to the nearest depth boundary. Across these methods, error peaks at the boundaries and drops to roughly one quarter of that level just one pixel away. \Cref{fig:teaser}(b) illustrates these errors for MonSter~\cite{MonSter} and IGEV~\cite{IGEV} on the same scene. In each column, the input left image and estimated disparity appear in the lower-right and upper-right insets, respectively, providing scene context for the main error map, where brighter colors indicate larger disparity errors. The green dashed boxes highlight the same region, where both methods exhibit bright error responses. These observations motivate our focus on  depth boundaries.

Photometric edges can respond to texture changes within a continuous
surface and therefore need not coincide with depth discontinuities.
For example, contours depicted in a picture on a planar surface can
produce strong image edges without corresponding changes in depth.
This motivates learning boundary cues from disparity structure.
We propose EG-Stereo, in which GeoEdgeNet predicts geometric edges
using labels derived from ground-truth disparity, without requiring
additional edge annotations.

Learning to locate boundaries alone does not specify how disparity
should be corrected.
EG-Stereo therefore uses the predicted geometric edge map in both
residual refinement and disparity supervision.
GeoRestore combines the map with the backbone disparity and
full-resolution image details to estimate disparity corrections,
while edge-weighted supervision increases error penalties at the
predicted boundaries.
The same spatial cue thus informs both the correction module and the
training objective.
Geometric edge supervision is retained during joint optimization,
while restoration supervision also updates the edge predictor.

On the Scene Flow dataset, EG-Stereo reduces EPE at depth boundaries by 13.4\%, 14.3\%, and 9.5\% for RAFT-Stereo~\cite{RAFT-Stereo}, IGEV, and MonSter, respectively, while  lowering their overall EPE from 0.72 to 0.67, 0.47 to 0.43, and 0.37 to 0.35 pixels. For IGEV, EPE at depth boundaries decreases from 3.77 to 3.23 pixels. Under the same training and evaluation protocols, EG-Stereo underperforms IGEV in zero-shot evaluation but outperforms it after fine-tuning. The fine-tuned model improves upon IGEV on all fourteen reported metrics across ETH3D, Middlebury 2014, KITTI 2012, and KITTI 2015. For example, ETH3D Bad1.0 (Noc) decreases from 1.12\% to 0.48\%, and KITTI 2015 D1-all (All) decreases from 1.59\% to 1.53\%. 

Our contributions are twofold.
We quantify how disparity errors concentrate near depth boundaries
across existing stereo methods.
We develop a framework that uses predicted geometric edges to condition
residual disparity correction and weight disparity supervision, and
demonstrate consistent boundary and overall accuracy gains across
three stereo backbones on Scene Flow.

%% file: sec/2_related.tex
\section{Related Work}\label{sec:related}

Recent stereo matching methods improve disparity estimation through cost-volume construction, iterative refinement, and auxiliary priors.

\noindent\textbf{Cost-Volume-Based Stereo Matching.}
Cost-volume-based methods construct matching costs from learned image features and aggregate them to estimate disparity~\cite{PSMNet,GwcNet,GANet,ACVNet,CoEx,LightStereo}. Representative approaches include PSMNet~\cite{PSMNet}, which captures multi-scale context through spatial pyramid pooling; GwcNet~\cite{GwcNet}, which constructs group-wise correlation volumes; and ACVNet~\cite{ACVNet}, which uses attention to filter matching information. Coarse-to-fine methods further reduce computation by progressively narrowing the disparity search range~\cite{Cas,CFNet}. Related transformer-based approaches model dense correspondences through attention~\cite{STTR,CSTR,GMStereo}. Other aggregation and correlation designs exploit local structure to improve matching and detail recovery, including adaptive aggregation in AANet~\cite{AANet}, bilateral aggregation in BANet~\cite{BANet}, and motif channel attention in MoCha-Stereo~\cite{MoCha-Stereo}. We use learned geometric edges to guide disparity refinement.

\noindent\textbf{Iterative Stereo Matching.}
Iterative methods progressively refine disparity estimates through repeated updates~\cite{RAFT-Stereo,IGEV,IGEV++,PCVNet,CREStereo,HITNet,MC-Stereo,CST-Stereo}. RAFT-Stereo~\cite{RAFT-Stereo} uses a recurrent update operator with multi-level correlation features, while IGEV and IGEV++~\cite{IGEV,IGEV++} incorporate geometry encoding volumes to improve initialization and subsequent updates. CREStereo~\cite{CREStereo} adopts cascaded recurrent refinement, and other methods improve matching representations or the update process~\cite{PCVNet,MC-Stereo,EAI-Stereo,CST-Stereo}. Selective-Stereo~\cite{Selective-Stereo} and DLNR~\cite{DLNR} further exploit frequency information to recover disparity details. A related optimization-based approach, NMRF-Stereo~\cite{NMRF-Stereo}, learns a Markov random field to preserve sharp disparity transitions. Our method uses predicted geometric edges for full-resolution disparity refinement and training.

\noindent\textbf{Prior-Guided Stereo Matching.}
Auxiliary priors provide complementary information for stereo matching in ambiguous regions. Monocular depth and pretrained visual representations have been incorporated to strengthen matching features and disparity estimation~\cite{MonSter,MonSter++,BridgeDepth,ViTAStereo,DEFOM-Stereo,StereoAnywhere,FoundationStereo}, while selective expert models improve robustness across diverse scenes~\cite{SMoEStereo}.  LoS~\cite{LoS} exploits local structure, and StereoNet~\cite{StereoNet} uses image-guided refinement to recover high-resolution disparity.  ADL~\cite{ADL} models multiple disparity modes at edge pixels during training. EdgeStereo~\cite{EdgeStereo} jointly models stereo matching and edge
detection, using photometric edge features and edge-aware regularization.
Our framework instead derives edge supervision from disparity structure
and uses the predicted geometric edge map to condition full-resolution
residual correction and weight disparity errors during training.

%% file: sec/3_method.tex
\begin{figure*}[t]
  \centering
  \includegraphics[width=0.96\linewidth]{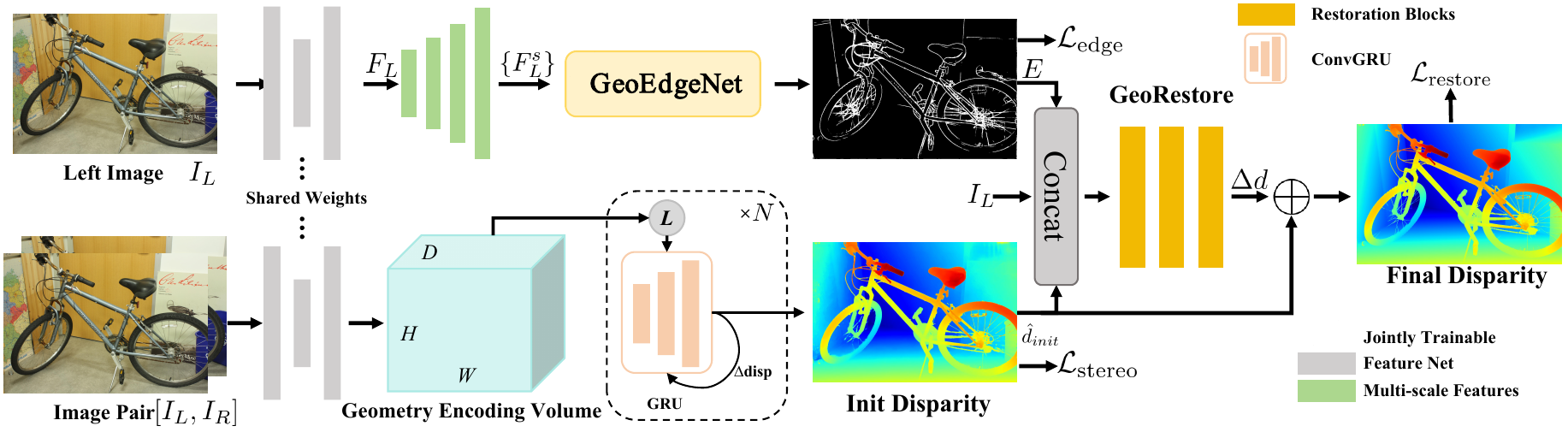}
  \caption{Overview of EG-Stereo. The Feature Net encodes the Image Pair, and the ConvGRU uses the Geometry Encoding Volume to estimate the Init Disparity. GeoEdgeNet predicts edges that guide GeoRestore to produce the Final Disparity.}
  \label{fig:framework}
\end{figure*}

\section{Method}\label{sec:method}
Improving disparity at depth boundaries requires identifying relevant boundaries, converting these cues into disparity corrections, and optimizing the cues for their downstream use.
As shown in \cref{fig:framework}, EG-Stereo addresses these requirements through a shared geometric edge map.
GeoEdgeNet learns boundary cues from disparity supervision, and GeoRestore combines them with the backbone disparity and image details to predict full-resolution residual corrections.
Joint training retains geometric edge supervision while using the same map to guide refinement and emphasize boundary errors.

\subsection{{GeoEdgeNet}}\label{sec:geoedge}

GeoEdgeNet targets boundaries associated with disparity structure rather than all visible image contours.
Appearance changes within a continuous surface can produce image edges that are irrelevant to depth discontinuities.
We therefore derive the supervision target from ground-truth disparity $d_{\mathrm{gt}}\in\mathbb{R}^{H\times W}$, defining the binary geometric edge map $Y_{\mathrm{edge}}\in\{0,1\}^{H\times W}$ as
\begin{equation}
\label{eq:edge_label}
  Y_{\mathrm{edge}}(x,y)
  =\mathbf{1}\bigl[
    \lvert\nabla^2 d_{\mathrm{gt}}(x,y)\rvert>1.0
  \bigr],
\end{equation}
where $\nabla^2$ measures second-order disparity changes and $\mathbf{1}[\cdot]$ is the indicator function.
These labels approximate depth boundaries from disparity structure without additional edge annotations.

\begin{figure}[t]
  \centering
  \includegraphics[width=0.95\linewidth]{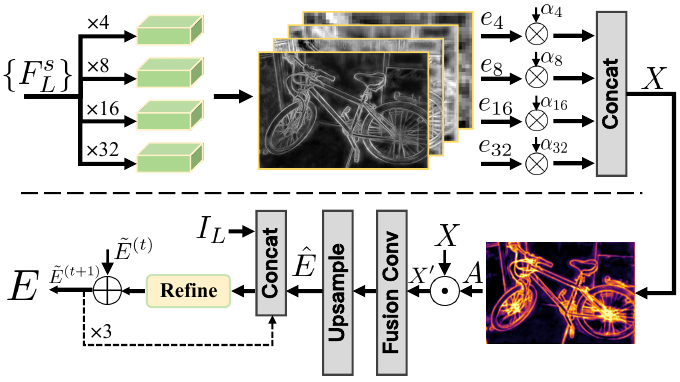}
  \caption{Architecture of GeoEdgeNet. }
  \label{fig:geoedgenet}
\end{figure}

As shown in \cref{fig:geoedgenet}, GeoEdgeNet predicts this target from shared backbone features and the left image.
It combines responses across feature scales and then uses full-resolution image detail to refine their spatial localization.

The stereo backbone applies a shared image encoder to $I_L,I_R\in\mathbb{R}^{3\times H\times W}$, where $H,W$ denote image height and width, as shown in \cref{fig:framework}. GeoEdgeNet reuses the left-image features $F_L=\{F_L^s\}_{s\in S}$, where $S=\{4,8,16,32\}$ denotes the spatial reduction factors. Each feature tensor $F_L^s\in\mathbb{R}^{C_s\times H/s\times W/s}$ has $C_s$ channels, determined by the stereo backbone.

To combine boundary cues from different feature scales, each $F_L^s$, $s\in S$, passes through two  convolutional layers to produce single-channel edge logits, which are bilinearly upsampled to $H/4\times W/4$. The resulting scale-wise edge logits are denoted by $e_s\in\mathbb{R}^{H/4\times W/4}$. Each $e_s$ is multiplied by a learnable scalar coefficient $\alpha_s$, and the four weighted maps are concatenated along the channel dimension:
\begin{equation}
X=\bigl[\alpha_s e_s\bigr]_{s\in S}
\in\mathbb{R}^{4\times H/4\times W/4}.
\end{equation}
Here, $[\cdot]$ denotes channel-wise concatenation.

The coefficients $\alpha_s$ control the contribution of each feature scale. A spatial attention module complements this scale weighting with location-dependent weights $A\in\mathbb{R}^{H/4\times W/4}$ predicted from $X$. A convolutional fusion layer $\Psi$ then combines the weighted responses into one logit map, and bilinear upsampling $U$ restores the original image size:
\begin{equation}
X'=A\odot X,\qquad
\hat E=U\bigl(\Psi(X')\bigr)
\in\mathbb{R}^{H\times W},
\end{equation}
where $\odot$ denotes element-wise multiplication with $A$ shared across channels.

To refine boundary localization with full-resolution image detail, we apply three residual updates conditioned on $I_L$. Starting from $\tilde E^{(0)}=\hat E$, each update concatenates the current edge logits $\tilde E^{(n-1)}$ with $I_L$ to form a four-channel tensor in $\mathbb{R}^{4\times H\times W}$. The convolutional edge update module $R$ predicts a single-channel residual, which is added to $\tilde E^{(n-1)}$. A sigmoid function $\sigma$ converts the logits after the third update into the geometric edge prediction:
\begin{equation}
\begin{aligned}
\tilde E^{(n)}
&=\tilde E^{(n-1)}
+R\bigl([\tilde E^{(n-1)},I_L]\bigr),
\quad n=1,2,3,\\
E&=\sigma\bigl(\tilde E^{(3)}\bigr)
\in[0,1]^{H\times W}.
\end{aligned}
\end{equation}

Each value of $E$ represents a predicted geometric edge probability.
Ground-truth disparity defines the supervision target during training,
while GeoEdgeNet requires only the left image and its shared backbone
features at inference.

\subsection{{GeoRestore}}\label{sec:refine}


A geometric edge map indicates potential depth boundaries but does not specify the direction or magnitude of disparity corrections.
GeoRestore addresses this gap by conditioning the correction on both the boundary cues and the disparity estimate to be refined.
As shown in \cref{fig:framework}, it receives the backbone disparity after the final recurrent update and upsampling, denoted by $\hat d_{\mathrm{init}}\in\mathbb{R}^{H\times W}$.

The three inputs provide complementary information.
The backbone disparity supplies the current local geometry, $E$ indicates potential depth boundaries, and $I_L$ provides full-resolution appearance details for localizing disparity transitions.
We concatenate $[\hat d_{\mathrm{init}}/d_{\mathrm{max}},I_L,E]$ into a tensor in $\mathbb{R}^{5\times H\times W}$, where $d_{\mathrm{max}}>0$ is the maximum disparity.

The correction module $\Phi$ converts these inputs into a residual $\Delta d$ in the original disparity units.
GeoRestore retains the backbone estimate as the base and adds the predicted residual to obtain the final disparity
\begin{equation}
\label{eq:georestore}
\begin{aligned}
  \Delta d
  &=\Phi\bigl(
    [\hat{d}_{\mathrm{init}}/d_{\mathrm{max}},\,I_L,\,E]
    \bigr)
    \in\mathbb{R}^{H\times W},\\
  \hat{d}_{\mathrm{final}}
  &=\hat{d}_{\mathrm{init}}+\Delta d
    \in\mathbb{R}^{H\times W}.
\end{aligned}
\end{equation}
Here, $[\cdot]$ denotes channel-wise concatenation. GeoRestore is applied once after the backbone's final update, and $\hat d_{\mathrm{final}}$ is the final disparity estimate of EG-Stereo.

We implement $\Phi$ with five convolutional layers with successive dilation rates $1,2,4,8,1$ to aggregate neighborhood information while preserving spatial resolution. Each layer is followed by instance normalization and a leaky ReLU activation. A final $1\times1$ convolution produces the single-channel residual. This  convolution is zero-initialized, and refinement starts with $\Delta d=0$ and $\hat d_{\mathrm{final}}=\hat d_{\mathrm{init}}$.

\subsection{{Joint Training}}\label{sec:loss}
The shared edge map serves two objectives during training.
Geometric edge supervision trains it to identify disparity-derived
boundaries, while restoration supervision updates it through its use
in disparity correction.
We jointly optimize GeoEdgeNet, the stereo backbone, and GeoRestore
under these objectives.
As shown in \cref{tab:joint_training}, the joint configuration reduces EPE$_{\mathrm{Edge}}$ from $3.53$ to $3.23$ compared with the frozen-predictor configuration.

\begin{table}[t]
\centering
\footnotesize
\setlength{\tabcolsep}{3pt}
\renewcommand{\arraystretch}{1.08}
\caption{Frozen-predictor and jointly trained configurations on Scene Flow. Results are reproduced from \cref{tab:ablation}.}
\label{tab:joint_training}
\begin{tabular*}{\columnwidth}{@{\extracolsep{\fill}}lccc@{}}
\toprule
Configuration
& EPE $\downarrow$
& EPE$_{\mathrm{Edge}}$ $\downarrow$
& Params (M) \\
\midrule
Pretrained, frozen GeoEdgeNet
& 0.46 & 3.53 & 14.90 \\
Jointly trained full model
& \textbf{0.43} & \textbf{3.23} & 12.78 \\
\bottomrule
\end{tabular*}
\end{table}

The predicted map $E$ connects the two tasks by conditioning GeoRestore and weighting disparity errors.
Let $\{d_i\}_{i=0}^{N}$, with $d_i\in\mathbb{R}^{H\times W}$, denote the stereo backbone's disparity predictions, where $N$ is the number of recurrent updates.
The prediction $d_0$ precedes these updates, and $d_N=\hat d_{\mathrm{init}}$ is supplied to GeoRestore.

Geometric edge supervision anchors $E$ to the disparity-derived labels in \cref{eq:edge_label}.
We supervise GeoEdgeNet's final edge logits with
\begin{equation}
\label{eq:edge_loss}
  \mathcal{L}_{\mathrm{edge}}
  =\operatorname{BCE}_{\mathrm{logits}}
  \bigl(\tilde{E}^{(3)},Y_{\mathrm{edge}}\bigr),
\end{equation}
where $\operatorname{BCE}_{\mathrm{logits}}$ denotes binary cross-entropy evaluated from logits.
For sparse ground truth, missing disparity values are filled before label construction, and edge supervision is restricted to originally annotated pixels.

To focus disparity training on predicted boundaries, we use the pixel weights $W_{\mathrm{pixel}}=1+3\,\mathrm{sg}(E)$.
The stop-gradient operator $\mathrm{sg}$ blocks gradients to $E$ through this weighting path.
The stereo disparity loss $\mathcal{L}_{\mathrm{stereo}}$ applies these weights to pixel-wise absolute errors of $\{d_i\}_{i=0}^{N}$.
The prediction $d_0$ has a sequence weight of $1$, while $d_i$ after update $i$ has weight $\gamma^{N-i}$ for $i\in\{1,\ldots,N\}$, where $\gamma\in(0,1)$ is a fixed decay coefficient.

GeoRestore's corrected output receives edge-weighted disparity supervision with a larger edge coefficient of $6$,
\begin{equation}
\mathcal{L}_{\mathrm{restore}}
=
\left\|
(1+6E)\odot
(\hat d_{\mathrm{final}}-d_{\mathrm{gt}})
\right\|_1,
\end{equation}
where $\odot$ denotes element-wise multiplication.
This coefficient increases the relative emphasis on boundary errors during refinement.
Gradients reach GeoEdgeNet through its input to GeoRestore and are also retained through $E$ in the restoration weight.
Both disparity supervision terms are evaluated only at pixels with valid ground truth.

The total objective is
$\mathcal{L}_{\mathrm{total}}
=
\mathcal{L}_{\mathrm{stereo}}
+0.5\mathcal{L}_{\mathrm{edge}}
+0.1\mathcal{L}_{\mathrm{restore}}$.
It jointly optimizes the shared image encoder, stereo backbone, GeoEdgeNet, and GeoRestore.
Geometric edge supervision retains the boundary prediction objective, while restoration supervision allows the edge predictor to adapt to disparity correction.
Ground-truth disparity and loss computation are required only during training.

%% file: sec/4_experiments.tex
\section{Experiments}\label{sec:experiments}

\subsection{Experimental Setup}\label{sec:setup}

\begin{table*}[t]
\centering
\setlength{\tabcolsep}{3.0pt}
\renewcommand{\arraystretch}{0.96}
\footnotesize
\caption{\textbf{Results on four real-world datasets.} Error rates are reported as percentages, RMSE in pixels, and runtime in milliseconds. ``All'' denotes all evaluated pixels, while ``Noc'' denotes non-occluded pixels. Results for competing methods are taken from official leaderboards or the corresponding publications. $^{\dagger}$ marks methods using large-scale pre-trained foundation models, which are included for reference. Among methods without a foundation prior, the \colorbox{firstbg}{best} and \colorbox{secondbg}{second-best} results are highlighted in red and yellow, respectively.}
\label{tab:real_benchmarks}
\begin{tabular}{l|ccc|ccc|cccc|cccc|c}
\toprule
\multirow{2}{*}{Method}
& \multicolumn{3}{c|}{ETH3D}
& \multicolumn{3}{c|}{Middlebury}
& \multicolumn{4}{c|}{KITTI 2015}
& \multicolumn{4}{c|}{KITTI 2012}
& \multirow{2}{*}{\shortstack{Runtime\\(ms) $\downarrow$}} \\
& \shortstack{Bad1.0\\Noc $\downarrow$} & \shortstack{Bad1.0\\All $\downarrow$} & \shortstack{RMSE\\Noc $\downarrow$}
& \shortstack{Bad2.0\\Noc $\downarrow$} & \shortstack{Bad2.0\\All $\downarrow$} & \shortstack{Bad4.0\\All $\downarrow$}
& \shortstack{D1-bg\\Noc $\downarrow$} & \shortstack{D1-all\\Noc $\downarrow$} & \shortstack{D1-bg\\All $\downarrow$} & \shortstack{D1-all\\All $\downarrow$}
& \shortstack{Out-2\\Noc $\downarrow$} & \shortstack{Out-2\\All $\downarrow$} & \shortstack{Out-3\\Noc $\downarrow$} & \shortstack{Out-3\\All $\downarrow$} & \\
\midrule
\multicolumn{16}{l}{\textit{Cost-volume methods}} \\
GwcNet~\cite{GwcNet} & 6.42 & 6.95 & 0.69 & -- & -- & -- & 1.61 & 1.92 & 1.74 & 2.11 & 2.16 & 2.71 & 1.32 & 1.70 & 320 \\
GANet~\cite{GANet} & 6.22 & 6.86 & 0.75 & -- & -- & -- & 1.34 & 1.63 & 1.48 & 1.81 & 1.89 & 2.50 & 1.19 & 1.60 & 1800 \\
AdaStereo~\cite{AdaStereo} & 3.09 & 3.34 & 0.44 & 13.7 & 19.8 & 10.9 & 2.39 & 2.83 & 2.59 & 3.08 & -- & -- & -- & -- & 410 \\
CasStereo~\cite{Cas} & -- & -- & -- & 18.8 & 26.0 & 18.5 & -- & -- & -- & -- & -- & -- & -- & -- & -- \\
LEAStereo~\cite{LEAStereo} & -- & -- & -- & 7.15 & 12.1 & 6.33 & 1.29 & 1.51 & 1.40 & 1.65 & 1.90 & 2.39 & 1.13 & 1.45 & 300 \\
ACVNet~\cite{ACVNet} & 2.58 & 2.86 & 0.45 & 13.6 & 19.5 & 12.6 & 1.26 & 1.52 & 1.37 & 1.65 & 1.83 & 2.35 & 1.13 & 1.47 & 200 \\
\midrule
\multicolumn{16}{l}{\textit{Iterative methods}} \\
RAFT-Stereo~\cite{RAFT-Stereo} & 2.44 & 2.60 & 0.36 & 4.74 & 9.37 & 6.42 & 1.44 & 1.69 & 1.58 & 1.82 & 1.92 & 2.42 & 1.30 & 1.66 & 380 \\
CREStereo~\cite{CREStereo} & 0.98 & 1.09 & \cellcolor{secondbg}0.28 & 3.71 & 8.13 & 5.05 & 1.33 & 1.54 & 1.45 & 1.69 & 1.72 & 2.18 & 1.14 & 1.46 & 410 \\
DLNR~\cite{DLNR} & -- & -- & -- & 3.20 & 6.98 & 4.77 & 1.42 & 1.61 & 1.60 & 1.76 & -- & -- & -- & -- & -- \\
Selective-IGEV~\cite{Selective-Stereo} & 1.23 & 1.56 & 0.29 & \cellcolor{firstbg}2.51 & \cellcolor{firstbg}6.04 & \cellcolor{firstbg}3.74 & 1.22 & \cellcolor{secondbg}1.44 & 1.33 & 1.55 & 1.59 & \cellcolor{secondbg}2.05 & 1.07 & 1.38 & 240 \\
LoS~\cite{LoS} & \cellcolor{secondbg}0.91 & \cellcolor{secondbg}1.03 & 0.31 & 4.20 & 8.03 & 5.07 & 1.29 & 1.52 & 1.42 & 1.65 & 1.69 & 2.12 & 1.10 & 1.38 & 190 \\
NMRF-Stereo~\cite{NMRF-Stereo} & -- & -- & -- & -- & -- & -- & \cellcolor{firstbg}1.18 & 1.46 & \cellcolor{firstbg}1.28 & 1.57 & 1.59 & 2.07 & \cellcolor{secondbg}1.01 & \cellcolor{secondbg}1.35 & \cellcolor{firstbg}90 \\
IGEV~\cite{IGEV} & 1.12 & 1.51 & 0.34 & 4.83 & 8.16 & 5.79 & 1.27 & 1.49 & 1.38 & 1.59 & 1.71 & 2.17 & 1.12 & 1.44 & \cellcolor{secondbg}180 \\
IGEV++~\cite{IGEV++} & 1.14 & 1.58 & 0.34 & 3.23 & 7.19 & 4.85 & 1.20 & \cellcolor{firstbg}1.42 & \cellcolor{secondbg}1.31 & \cellcolor{firstbg}1.51 & \cellcolor{firstbg}1.56 & \cellcolor{firstbg}2.03 & 1.04 & 1.36 & 280 \\
\midrule
\multicolumn{16}{l}{\textit{Edge-guided methods}} \\
EdgeStereo~\cite{EdgeStereo} & 6.76 & 7.12 & 0.81 & 18.7 & 24.5 & 15.7 & 1.69 & 1.89 & 1.84 & 2.08 & 2.32 & 2.88 & 1.46 & 1.83 & 320 \\
\midrule
\rowcolor[gray]{0.9}
EG-Stereo (Ours) & \cellcolor{firstbg}0.48 & \cellcolor{firstbg}0.82 & \cellcolor{firstbg}0.21 & \cellcolor{secondbg}2.95 & \cellcolor{secondbg}6.58 & \cellcolor{secondbg}4.12 & \cellcolor{secondbg}1.19 & \cellcolor{firstbg}1.42 & \cellcolor{secondbg}1.31 & \cellcolor{secondbg}1.53 & \cellcolor{secondbg}1.58 & \cellcolor{firstbg}2.03 & \cellcolor{firstbg}1.00 & \cellcolor{firstbg}1.32 & 190   \\
\midrule
\multicolumn{16}{l}{\textcolor{gray}{\textit{Foundation-model-based methods (for reference)}}} \\
\textcolor{gray}{CroCo-Stereo$^{\dagger}$~\cite{croco_v2}} & \textcolor{gray}{0.99} & \textcolor{gray}{1.14} & \textcolor{gray}{0.30} & \textcolor{gray}{7.29} & \textcolor{gray}{11.1} & \textcolor{gray}{6.75} & \textcolor{gray}{1.30} & \textcolor{gray}{1.51} & \textcolor{gray}{1.38} & \textcolor{gray}{1.59} & \textcolor{gray}{--} & \textcolor{gray}{--} & \textcolor{gray}{--} & \textcolor{gray}{--} & \textcolor{gray}{930} \\
\textcolor{gray}{MonSter$^{\dagger}$~\cite{MonSter}} & \textcolor{gray}{0.46} & \textcolor{gray}{0.72} & \textcolor{gray}{0.20} & \textcolor{gray}{2.64} & \textcolor{gray}{6.14} & \textcolor{gray}{--} & \textcolor{gray}{1.05} & \textcolor{gray}{1.33} & \textcolor{gray}{1.13} & \textcolor{gray}{1.41} & \textcolor{gray}{1.36} & \textcolor{gray}{1.75} & \textcolor{gray}{0.84} & \textcolor{gray}{1.09} & \textcolor{gray}{450} \\
\bottomrule
\end{tabular}
\end{table*}


\noindent\textbf{Datasets.}
We use Scene Flow~\cite{MIFDB16} for pre-training and evaluation, with 35,454 training stereo pairs and 4,370 test stereo pairs providing dense ground-truth disparity. The real-world datasets are ETH3D~\cite{schops2017eth3d}, Middlebury 2014~\cite{scharstein2014highres}, KITTI 2015~\cite{Menze2015CVPR}, and KITTI 2012~\cite{Geiger2012CVPR}.

\noindent\textbf{Metrics.}
On the Scene Flow dataset, we report end-point error (EPE) and EPE at depth boundaries ($\mathrm{EPE}_{\mathrm{Edge}}$). For fine-tuned test-set evaluation, we submit predicted disparity maps to the official evaluation servers and report the returned Bad1.0 and RMSE metrics for ETH3D, Bad2.0 and Bad4.0 for Middlebury 2014, D1-bg and D1-all for KITTI 2015, and Out-2 and Out-3 for KITTI 2012. Zero-shot errors are computed locally using thresholds of 3 pixels on KITTI, 2 pixels on Middlebury at quarter and half resolution, and 1 pixel on ETH3D.

To assess the quality of geometric edge guidance, we also evaluate edge predictions against the geometric edge labels using precision, recall, and F1, with a matching tolerance of one pixel. Precision and recall measure the fractions of predicted and labeled edge pixels that are matched, respectively, and F1 is their harmonic mean.
Zero-shot error rates are computed on the full real-world training sets. EPE and the per-image statistics use fixed random subsets of 50 images each from KITTI~2012 and KITTI~2015, five Middlebury~2014 images at half resolution, and 100 Scene Flow test images, together with all 27 ETH3D images.

\noindent\textbf{Implementation details.}
EG-Stereo uses the same geometric edge map for disparity refinement and loss weighting across RAFT-Stereo~\cite{RAFT-Stereo},
IGEV~\cite{IGEV}, and MonSter~\cite{MonSter}, reusing each backbone's
encoder features for GeoEdgeNet.
Pre-training uses 200k steps, batch size 8, $320\times736$ crops,
and 22/32 recurrent updates during training/inference.
We fix the coefficients of $\mathcal{L}_{\mathrm{edge}}$ and
$\mathcal{L}_{\mathrm{restore}}$ at 0.5 and 0.1,
the disparity edge weights at 3 and 6, and the geometric edge
label threshold at 1.0.
All experiments use eight NVIDIA RTX 5090 GPUs with mixed-precision
training.

IGEV is the reference backbone for zero-shot evaluation, fine-tuning,
and ablation.
Our implementation uses its official code and retains its baseline
Scene Flow training and inference settings specified above,
optimizer, learning-rate schedule, augmentation, and fine-tuning
data mixtures.

Fine-tuning starts from the IGEV-based Scene Flow checkpoint,
with unchanged loss coefficients and edge label threshold.
We train for 50k steps on the combined KITTI 2012 and KITTI 2015
training sets.
Following CREStereo and GMStereo, ETH3D fine-tuning uses 300k steps
on a mixture of TartanAir, CREStereo, Scene Flow, Sintel Stereo,
InStereo2k, and ETH3D.
Middlebury 2014 uses the same mixture with ETH3D replaced by
Middlebury 2014.
Geometric edge labels are derived from the training ground-truth
disparities.

\subsection{Comparison with Existing Methods}\label{sec:comparison}
\begin{figure}[t]
  \centering
  \includegraphics[width=0.9\linewidth]{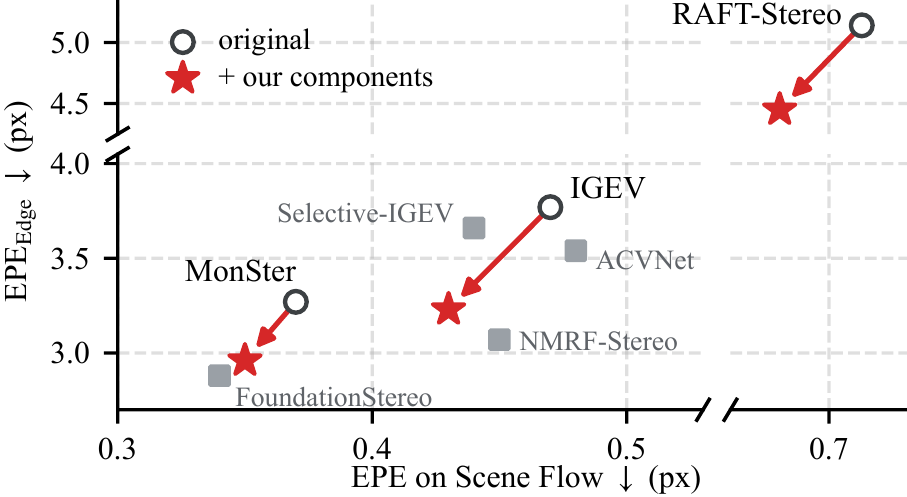}
  \caption{Overall EPE and boundary EPE$_{\mathrm{Edge}}$ on the Scene Flow test set before and after adding our components.}
  \label{fig:sceneflow_overview}
\end{figure}

\begin{table}[t]
\centering
\footnotesize
\setlength{\tabcolsep}{3pt}
\renewcommand{\arraystretch}{1.08}
\caption{Results of geometric edge guidance across three stereo backbones on Scene Flow.}
\label{tab:backbone_results}
\resizebox{\columnwidth}{!}{%
\begin{tabular}{@{}lcccc@{}}
\toprule
Backbone
& EPE $\downarrow$
& EPE$_{\mathrm{Edge}}$ $\downarrow$
& Params (M)
& Time (ms) \\
\midrule
RAFT-Stereo~\cite{RAFT-Stereo}
& 0.72\,$\rightarrow$\,\textbf{0.67}
& 5.14\,$\rightarrow$\,\textbf{4.45}
& 11.12\,$\rightarrow$\,11.30
& 380\,$\rightarrow$\,390 \\
IGEV~\cite{IGEV}
& 0.47\,$\rightarrow$\,\textbf{0.43}
& 3.77\,$\rightarrow$\,\textbf{3.23}
& 12.60\,$\rightarrow$\,12.78
& 180\,$\rightarrow$\,190 \\
MonSter~\cite{MonSter}
& 0.37\,$\rightarrow$\,\textbf{0.35}
& 3.27\,$\rightarrow$\,\textbf{2.96}
& 388.69\,$\rightarrow$\,388.87
& 450\,$\rightarrow$\,460 \\
\bottomrule
\end{tabular}%
}
\end{table}

\noindent\textbf{Pre-training Performance on Scene Flow.}
To evaluate whether EG-Stereo improves disparity estimation across different stereo matching networks, we incorporate its three components into RAFT-Stereo~\cite{RAFT-Stereo}, IGEV~\cite{IGEV}, and MonSter~\cite{MonSter}, and compare each implementation with its corresponding baseline on the Scene Flow test set. As shown in~\cref{tab:backbone_results}, our components reduce both EPE and EPE$_{\mathrm{Edge}}$ across all three networks. Each table entry gives the baseline value followed by the value with our components. EPE$_{\mathrm{Edge}}$ decreases from $5.14$ to $4.45$ for RAFT-Stereo, from $3.77$ to $3.23$ for IGEV, and from $3.27$ to $2.96$ for MonSter, corresponding to reductions of $13.4\%$, $14.3\%$, and $9.5\%$, respectively. Overall EPE also decreases from $0.72$ to $0.67$, from $0.47$ to $0.43$, and from $0.37$ to $0.35$, respectively. These consistent improvements support using learned geometric edges to guide disparity refinement and training across different stereo matching networks.

We further compare our implementations with existing methods, as shown in~\cref{fig:sceneflow_overview}. Our IGEV-based implementation achieves an EPE of $0.43$ and an EPE$_{\mathrm{Edge}}$ of $3.23$, compared with $0.44$ and $3.66$ for Selective-IGEV~\cite{Selective-Stereo}. Our best-performing implementation, built on MonSter, achieves an EPE of $0.35$ and an EPE$_{\mathrm{Edge}}$ of $2.96$, compared with $0.45$ and $3.07$ for NMRF-Stereo~\cite{NMRF-Stereo}. FoundationStereo~\cite{FoundationStereo} achieves lower errors on both metrics. The improvement over MonSter shows that our framework also benefits a foundation-model-based backbone.

\begin{table}[t]
\centering
\footnotesize
\renewcommand{\arraystretch}{1.1}
\caption{Zero-shot results and retraining with downstream datasets on real-world datasets.}
\label{tab:zeroshot}
\begin{tabular*}{\columnwidth}{@{\extracolsep{\fill}}lcccc@{}}
\toprule
\multirow{2}{*}{Dataset}
& \multicolumn{2}{c}{Zero-shot}
& \multicolumn{2}{c}{\shortstack{Retraining}} \\
\cmidrule(lr){2-3}\cmidrule(lr){4-5}
& IGEV & EG-Stereo & IGEV & EG-Stereo \\
\midrule
\shortstack[l]{KITTI 2012}   & \textbf{5.14} & 5.59 & 1.44 & \textbf{1.32} \\
\shortstack[l]{Middlebury 2014}   & \textbf{7.08} & 7.43 & 8.16 & \textbf{6.58} \\
\shortstack[l]{ETH3D}   & \textbf{3.61} & 3.97 & 1.51 & \textbf{0.82} \\
\bottomrule
\end{tabular*}
\end{table}

\noindent\textbf{Fine-Tuning Performance on Real-World Datasets.}
Using IGEV as the reference backbone without a foundation-model prior, we evaluate EG-Stereo after fine-tuning on real-world datasets.
\cref{tab:real_benchmarks} reports the official test results on ETH3D, Middlebury 2014, KITTI 2015, and KITTI 2012. EG-Stereo improves over its IGEV baseline on all fourteen reported metrics. Averaged equally across the fourteen metrics, the relative error reduction over the fine-tuned IGEV baseline is 20.1\%. Specifically, on ETH3D, Bad1.0 (Noc) decreases from 1.12\% to 0.48\%. On Middlebury 2014, Bad2.0 (Noc) decreases from 4.83\% to 2.95\%. The improvement also extends to KITTI, with D1-all (All) decreasing from 1.59\% to 1.53\% on KITTI 2015 and Out-3 (Noc) decreasing from 1.12\% to 1.00\% on KITTI 2012.   Although these official test evaluations do not report EPE$_{\mathrm{Edge}}$, the improvements show that using geometric edges to guide disparity refinement and training also improves overall disparity estimation after fine-tuning on real-world datasets.

Among methods without a foundation-model prior, our model outperforms Selective-IGEV~\cite{Selective-Stereo}  on all eleven reported ETH3D and KITTI metrics.
It also outperforms IGEV++~\cite{IGEV++} on all three ETH3D and all three Middlebury metrics, and matches or improves on it on six of the eight KITTI metrics.
MonSter~\cite{MonSter}, which uses a pre-trained foundation model,  is included for reference.

\noindent\textbf{Zero-Shot Performance on Real-World Datasets.}
\label{sec:zeroshot}
We evaluate the Scene Flow checkpoints of IGEV and EG-Stereo on the real-world training splits without target-domain training. As shown in \cref{tab:zeroshot}, EG-Stereo yields higher error rates than IGEV in all five zero-shot settings, with absolute increases of 0.21–0.59 percentage points. In the failure cases we inspect, EG-Stereo often mislocalizes depth boundaries. We attribute this in part to differences in boundary geometry between the synthetic pretraining data and the real-world datasets. In the examples we examine, synthetic depth boundaries are often straight and aligned with the image axes, whereas real-world boundaries can be irregular or jagged.

To test this explanation, we retrain the models with samples representative of the downstream datasets. After retraining, EG-Stereo reduces Out-3 on KITTI 2012 from 1.44\% to 1.32\%, Bad-2 on Middlebury 2014 from 8.16\% to 6.58\%, and Bad-1 on ETH3D from 1.51\% to 0.82\%, relative to IGEV. These results support our explanation that differences in depth-boundary geometry between synthetic and real-world data contribute to the zero-shot degradation.

\begin{table}[t]
\centering
\footnotesize
\setlength{\tabcolsep}{3pt}
\renewcommand{\arraystretch}{1.08}
\caption{Ablation results on the Scene Flow test set.}
\label{tab:ablation}
\resizebox{\columnwidth}{!}{%
\begin{tabular}{@{}lcccc@{}}
\toprule
Configuration
& EPE $\downarrow$
& EPE$_{\mathrm{Edge}}$ $\downarrow$
& Params (M)
& Time (ms) \\
\midrule
\multicolumn{5}{@{}l}{\textit{Progressive component addition}} \\
IGEV baseline
& 0.47 & 3.77 & 12.60 & 180 \\
+ edge-weighted loss (GT edges)
& 0.46 & 3.68 & 12.60 & 180 \\
+ GeoEdgeNet
& 0.45 & 3.42 & 12.74 & 186 \\
\rowcolor[gray]{0.9}
+ GeoRestore
& 0.43 & 3.23 & 12.78 & 190 \\
\midrule
\multicolumn{5}{@{}l}{\textit{Additional ablations}} \\
GeoEdgeNet only
& 0.46 & 3.65 & 12.74 & 186 \\
No edge input
& 0.46 & 3.81 & 12.78 & 190 \\
Fixed GeoEdgeNet
& 0.46 & 3.53 & 14.90 & 190 \\
\bottomrule
\end{tabular}%
}
\end{table}

\begin{table}[!ht]
\centering
\small
\setlength{\tabcolsep}{4pt}
\renewcommand{\arraystretch}{1.08}
\caption{Edge prediction quality, disparity accuracy, and computational cost are compared across edge sources on Scene Flow.}
\label{tab:edge_quality}
\begin{adjustbox}{max width=\linewidth}
\begin{tabular}{@{}lccccccc@{}}
\toprule
Edge source & Precision $\uparrow$ & Recall $\uparrow$ & F1 $\uparrow$
& EPE $\downarrow$ & EPE$_{\mathrm{Edge}}$ $\downarrow$ & Params (M) & Time (ms) \\
\midrule
Sobel            & 0.456 & 0.627 & 0.528 & 0.49 & 3.86 & 12.64 & 182 \\
Canny            & 0.456 & 0.589 & 0.514 & 0.47 & 3.79 & 12.64 & 183 \\
Frozen RCF~\cite{RCF} & 0.633 & 0.729 & 0.677 & 0.51 & 4.12 & 27.44 & 193 \\
\rowcolor[gray]{0.9}
GeoEdgeNet       & 0.818 & 0.847 & 0.832 & 0.43 & 3.23 & 12.78 & 190 \\
\bottomrule
\end{tabular}
\end{adjustbox}
\end{table}

\subsection{Ablation Study}\label{sec:ablation}

\noindent\textbf{Contribution of each component.}
We examine how geometric edge prediction contributes to disparity
estimation through auxiliary supervision, disparity weighting,
and residual correction.
On the Scene Flow test set, \cref{tab:ablation} first compares successive
IGEV-based configurations with ground-truth edge weighting,
predicted edge weighting, and GeoRestore, followed by additional
ablations of how the edge predictions are used.
First, applying the edge-weighted loss with weights constructed from ground-truth geometric edge labels reduces EPE$_{\mathrm{Edge}}$ from $3.77$ to $3.68$. Adding GeoEdgeNet and using its predicted geometric edges for weighting further reduces EPE$_{\mathrm{Edge}}$ to $3.42$. Finally, adding GeoRestore reduces EPE$_{\mathrm{Edge}}$ to $3.23$ and overall EPE from $0.45$ to $0.43$.

\noindent\textbf{Auxiliary prediction versus explicit edge use.}
``GeoEdgeNet only'' adds auxiliary edge prediction to IGEV without using the predictions for disparity weighting or refinement.
Its EPE$_{\mathrm{Edge}}$ is $3.65$, compared with $3.77$ for IGEV and $3.42$ when predicted edges also weight disparity errors.
In this comparison, auxiliary edge prediction alone provides a smaller
gain than also using its predictions to weight disparity errors.
The benefit therefore extends beyond adding an edge prediction task
to the shared backbone.

\noindent\textbf{Edge guidance for disparity correction.}
In the ``No edge input'' configuration, the geometric edge input to GeoRestore is set to zero.
EPE$_{\mathrm{Edge}}$ increases from $3.23$ to $3.81$, supporting the role of geometric edge input in the evaluated refinement configuration.

\noindent\textbf{Frozen versus jointly trained edge prediction.}
``Fixed GeoEdgeNet'' uses a separately pretrained and frozen edge predictor.
It yields an EPE of $0.46$ and an EPE$_{\mathrm{Edge}}$ of $3.53$, compared with $0.43$ and $3.23$ for the jointly trained full model.
The joint configuration therefore provides more effective guidance for disparity estimation in this comparison.

\subsection{Depth Boundary Analysis}

\noindent\textbf{Comparison of edge sources.}
To assess how well different edge sources identify geometric boundaries and support disparity estimation, we compare GeoEdgeNet with Sobel~\cite{Sobel1968}, Canny~\cite{Canny1986}, and frozen RCF~\cite{RCF}. Sobel detects local image intensity changes using fixed gradient filters. Canny suppresses noisy responses and links thin edges into contours. RCF is a pre-trained neural network for image edge detection whose weights are kept fixed in this comparison. We evaluate precision, recall, and F1 against reference geometric edges with a matching tolerance of one pixel. Precision measures the fraction of predicted edge pixels matched to reference edges, recall measures the fraction of reference edge pixels recovered, and F1 balances the two. Together, these metrics quantify unwanted edge responses and missed geometric boundaries, complementing the disparity errors. Parameter counts and runtimes refer to the complete stereo models, with the GeoEdgeNet row corresponding to EG-Stereo.
\begin{figure}[t]
  \centering
  \includegraphics[width=\linewidth]{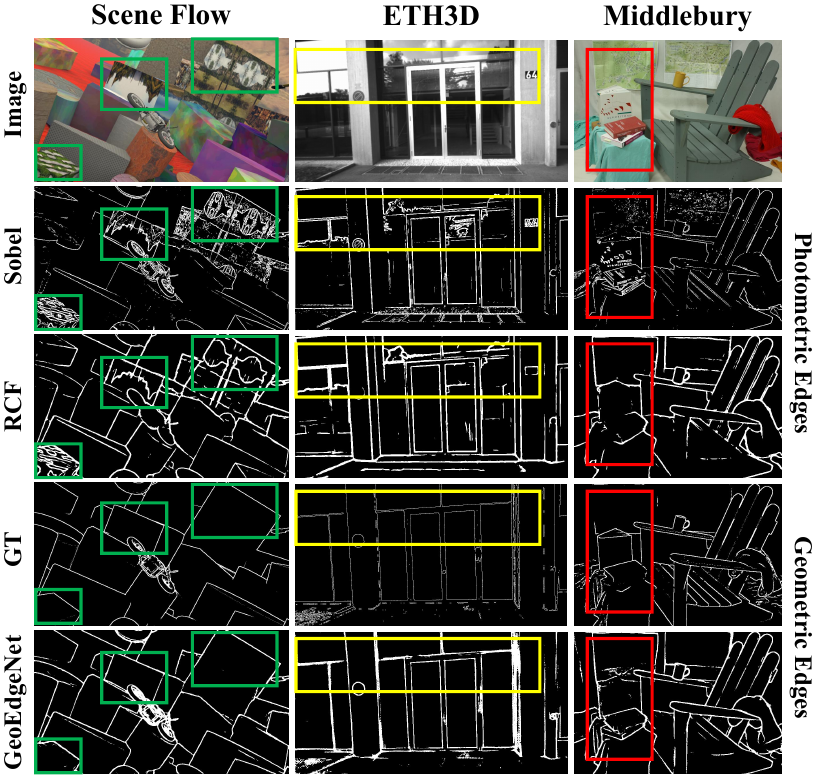}
  \caption{Qualitative comparison of edge sources on Scene Flow, ETH3D, and Middlebury. }
  \label{fig:geometric_edge}
\end{figure}

\begin{figure*}[!ht]
  \centering
  \includegraphics[width=0.99\linewidth]{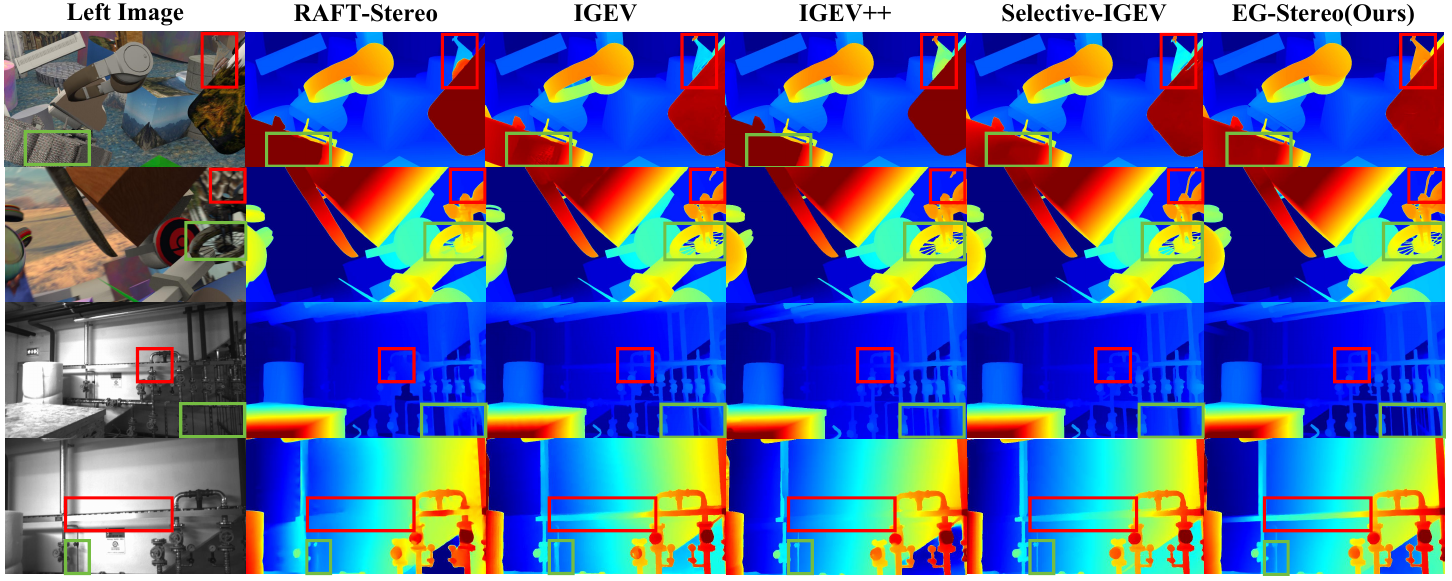}
  \caption{Qualitative disparity comparisons on Scene Flow and ETH3D.  The first two rows show a foreground structure and wheel spokes from Scene Flow. The last two rows show railing bars and a horizontal pipe from ETH3D. Colored boxes highlight regions used to compare depth boundaries and thin structures.}
  \label{fig:qualitative}
\end{figure*}

As shown in \cref{tab:edge_quality}, GeoEdgeNet achieves the highest precision, recall, and F1, together with the lowest overall and boundary disparity errors.
Its precision is $0.818$, compared with $0.633$ for RCF, indicating a smaller fraction of predictions unmatched to the reference geometric boundaries.
Its F1 increases to $0.832$ from RCF's $0.677$, while EPE$_{\mathrm{Edge}}$ is $3.23$, compared with $3.79$ for Canny and $4.12$ for RCF.
RCF has higher F1 than Canny but larger boundary disparity error,
showing that edge F1 alone is insufficient to assess the usefulness
of disparity guidance.
The comparison therefore evaluates both agreement with geometric
boundaries and the resulting disparity accuracy.
GeoEdgeNet achieves the best results on both aspects among the
evaluated configurations.

\cref{fig:geometric_edge} illustrates the distinction between appearance contours and geometric boundaries.
In the green boxes of the Scene Flow example, Sobel and RCF respond to contours within pictures on planar surfaces, including contours absent from the reference geometric edge map.
These responses describe image content rather than corresponding depth discontinuities.
GeoEdgeNet suppresses several such internal responses while retaining the surrounding geometric boundaries, although some appearance responses remain.
This example illustrates why boundary guidance should be evaluated against disparity structure rather than image contours alone.

\noindent\textbf{Error versus distance to depth boundaries.}

\begin{wrapfigure}{l}{0.48\columnwidth}
  \centering
  \includegraphics[width=\linewidth]{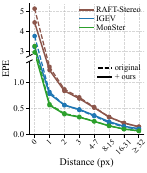}
  \caption{EPE versus distance to depth boundaries across three stereo backbones.}
  \label{fig:dist}
\end{wrapfigure}

To examine where our components improve disparity estimates, we group Scene Flow test pixels by their distance to the nearest depth boundary. As shown in~\cref{fig:dist}, we compare RAFT-Stereo, IGEV, and MonSter with their corresponding EG-Stereo implementations. Distance zero denotes the geometric edge pixels. Our components reduce EPE at these pixels across all three backbones.

For IGEV, geometric edge pixels occupy only $6.0\%$ of the evaluated pixels but contribute about half of the total absolute disparity error. Adding our components reduces EPE at these pixels from $3.77$ to $3.23$, corresponding to a $14.3\%$ reduction, and also reduces EPE by $4.2\%$ one pixel away. Farther from the boundaries, some distance groups have small EPE increases of up to $0.02$ pixels, while overall EPE decreases from $0.47$ to $0.43$. This analysis shows that the improvement  is concentrated near depth boundaries and contributes to better overall disparity accuracy.

\noindent\textbf{Qualitative results.}
To inspect disparity estimates near depth boundaries, we compare the IGEV implementation of EG-Stereo with RAFT-Stereo, IGEV, IGEV++, and Selective-IGEV on examples from Scene Flow and ETH3D.
As shown in \cref{fig:qualitative}, the highlighted regions reveal differences in object separation and the completeness of thin structures.
On Scene Flow, EG-Stereo represents the foreground structure in the first-row red box more completely than IGEV and separates the wheel spokes in the second-row green box more clearly than RAFT-Stereo.
On ETH3D, the fine-tuned model preserves more individual railing bars
in the third-row green box and produces a more continuous horizontal
pipe in the fourth-row red box than IGEV.
These examples complement the distance analysis by illustrating clearer boundaries and more complete thin structures in the displayed disparity estimates.

%% file: sec/5_conclusion.tex
\section{Conclusion}

We found that disparity errors on Scene Flow peak at depth boundaries across the stereo methods and drop sharply just one pixel away. To improve disparity estimates at these boundaries, EG-Stereo learns a geometric edge map from ground-truth disparity. The predicted map guides full-resolution residual refinement and gives greater weight to disparity errors during training. On Scene Flow, EG-Stereo reduces boundary EPE by $9.5\%$--$14.3\%$ across RAFT-Stereo, IGEV, and MonSter, while also reducing overall EPE.
After fine-tuning, its IGEV-based implementation improves upon IGEV on all fourteen reported metrics across ETH3D, Middlebury 2014, KITTI 2012, and KITTI 2015.


\section*{LLM Usage Statement}
OpenAI ChatGPT was used to assist with language editing and proofreading
of the manuscript.